%% file: parseval.tex
\documentclass{article}

\usepackage{times}
\usepackage{graphicx} 
\usepackage{tabularx}
\usepackage{multirow}

\usepackage{hyperref}
\usepackage{xfunctions}
\usepackage[utf8]{inputenc} 
\usepackage[T1]{fontenc}    
\usepackage{hyperref}       
\usepackage{url}            
\usepackage{booktabs}       
\usepackage{amsfonts}       
\usepackage{amsmath}
\usepackage{amsthm}
\usepackage{amssymb}
\usepackage{graphicx}
\usepackage{caption}
\usepackage{subcaption}
\usepackage{bbm}
\usepackage[ruled,vlined]{algorithm2e}

\include{macros}

\newtheorem{theorem}{Theorem}

\newtheorem{remark}[theorem]{Remark}

\newcommand{\myparagraph}[1]{\textbf{#1}}

\usepackage[accepted]{icml2017}

\icmltitlerunning{Parseval Networks}

\begin{document}

\twocolumn[
\icmltitle{Parseval Networks: Improving Robustness to Adversarial Examples}

\begin{icmlauthorlist}
\icmlauthor{Moustapha Cisse}{to}
\icmlauthor{Piotr Bojanowski}{to}
\icmlauthor{Edouard Grave}{to}
\icmlauthor{Yann Dauphin}{to}
\icmlauthor{Nicolas Usunier}{to}
\end{icmlauthorlist}

\icmlaffiliation{to}{Facebook AI Research}

\icmlcorrespondingauthor{Moustapha Cisse}{moustaphacisse@fb.com}
\icmlkeywords{deep learning, security, robustness, ICML}

\vskip 0.3in
]
\printAffiliationsAndNotice{}
\begin{abstract}

  We introduce Parseval networks, a form of deep neural networks in
  which the Lipschitz constant of linear, convolutional and
  aggregation layers is constrained to be smaller than $1$. Parseval
  networks are empirically and theoretically motivated by an analysis
  of the robustness of the predictions made by deep neural networks
  when their input is subject to an adversarial perturbation. The most
  important feature of Parseval networks is to maintain weight
  matrices of linear and convolutional layers to be (approximately)
  Parseval tight frames, which are extensions of orthogonal matrices
  to non-square matrices. We describe how these constraints can be
  maintained efficiently during SGD. We show that Parseval networks
  match the state-of-the-art in terms of accuracy on CIFAR-10/100 and
  Street View House Numbers (SVHN), while being more robust than their
  vanilla counterpart against adversarial examples.  Incidentally,
  Parseval networks also tend to train faster and make a better usage
  of the full capacity of the networks.

\end{abstract}

\section{Introduction}

Deep neural networks achieve near-human accuracy on many perception
tasks \cite{he2016deep,amodei2015deep}. However, they lack robustness
to small alterations of the inputs at test
time~\cite{szegedy2013intriguing}. Indeed when presented with a
corrupted image that is barely distinguishable from a legitimate one
by a human, they can predict incorrect labels, with
high-confidence. An adversary can design such so-called adversarial
examples, by adding a small perturbation to a legitimate input to
maximize the likelihood of an incorrect class under constraints on the
magnitude of the
perturbation~\cite{szegedy2013intriguing,goodfellow2015explaining,moosavi2015deepfool,papernot2016practical}.
In practice, for a significant portion of inputs, a single step in the
direction of the gradient sign is sufficient to generate an
adversarial example~\cite{goodfellow2015explaining} that is even
transferable from one network to another one trained for the same
problem but with a different architecture \cite{DBLP:journals/corr/LiuCLS16,kurakin2016adversarial}.

The existence of transferable adversarial examples has two undesirable corollaries. First, it creates a security threat for production systems by enabling black-box attacks~\cite{papernot2016practical}. Second, it underlines the lack of \emph{robustness} of neural networks and questions their ability to generalize in settings where the train and test distributions can be (slightly) different as is the case for the distributions of legitimate and adversarial examples.

Whereas the earliest works on adversarial examples already suggested
that their existence was related to the magnitude of the hidden
activations gradient with respect to their
inputs~\cite{szegedy2013intriguing}, they also empirically assessed
that standard regularization schemes such as weight decay or training
with random noise do not solve the
problem~\cite{goodfellow2015explaining,fawzi2016robustness}. The
current mainstream approach to improving the robustness of deep
networks is adversarial training. It consists in generating
adversarial examples on-line using the current network's
parameters~\cite{goodfellow2015explaining,miyato2015distributional,moosavi2015deepfool,szegedy2013intriguing,kurakin2016adversarial}
and adding them to the training data. This data augmentation method
can be interpreted as a robust optimization
procedure~\cite{shaham2015understanding}.

In this paper, we introduce Parseval networks, a layerwise
regularization method for reducing the network's sensitivity to small
perturbations by carefully controlling its global Lipschitz constant.
Since the network is a composition of functions represented by its
layers, we achieve increased robustness by maintaining a small
Lipschitz constant (e.g., 1) at every hidden layer; be it
fully-connected, convolutional or residual. In particular, a critical
quantity governing the local Lipschitz constant in both fully
connected and convolutional layers is the spectral norm of the weight
matrix. Our main idea is to control this norm by parameterizing the
network with \emph{parseval tight
  frames}~\cite{kovavcevic2008introduction}, a generalization of
orthogonal matrices.

The idea that regularizing the spectral norm of each weight matrix
could help in the context of robustness appeared as early as
\cite{szegedy2013intriguing}, but no experiment nor algorithm was
proposed, and no clear conclusion was drawn on how to deal with
convolutional layers. Previous work, such as double backpropagation~\cite{drucker1992improving} has also explored jacobian normalization as a way to improve generalization.
Our contribution is twofold. First, we provide
a deeper analysis which applies to fully connected networks, convolutional networks, as well as Residual networks \cite{he2016deep}.
Second, we propose a computationally efficient algorithm and validate its effectiveness on standard benchmark datasets.
We report results on MNIST, CIFAR-10, CIFAR-100
and Street View House Numbers (SVHN), in which fully connected and wide residual networks were trained \cite{zagoruyko2016wide} with
Parseval regularization. The accuracy of Parseval networks on legitimate test examples matches the
state-of-the-art, while the results show notable improvements on
adversarial examples. Besides, Parseval networks train significantly faster than their vanilla counterpart.

In the remainder of the paper, we first discuss the previous work on adversarial examples. Next, we give formal definitions of the
adversarial examples and provide an analysis of the robustness of deep neural networks. Then, we introduce Parseval networks and its efficient training algorithm. Section 5 presents experimental results validating the model and providing several insights.

\section{Related work}

Early papers on adversarial examples attributed the vulnerability of
deep networks to high local
variations~\cite{szegedy2013intriguing,goodfellow2015explaining}. Some
authors argued that this sensitivity of deep networks to small changes
in their inputs is because neural networks only learn the
discriminative information sufficient to obtain good accuracy rather
than capturing the true concepts defining the
classes~\cite{fawzi2015analysis, nguyen2015deep}.

Strategies to improve the robustness of deep networks include
defensive distillation~\cite{papernot2016distillation}, as well as
various regularization procedures such as contractive
networks~\cite{gu2014towards}. However, the bulk of recent proposals
relies on data augmentation
\cite{goodfellow2015explaining,miyato2015distributional,moosavi2015deepfool,shaham2015understanding,szegedy2013intriguing,kurakin2016adversarial}. It
uses adversarial examples generated online during training. As we
shall see in the experimental section, regularization can be
complemented with data augmentation; in particular, Parseval networks
with data augmentation appear more robust than either data
augmentation or Parseval networks considered in isolation.

\section{Robustness in Neural Networks}

We consider a multiclass prediction setting, where we have $\nC$
classes in $\Cset = \{1, ..., \nC\}$. A multiclass classifier is a
function $\cnet:(\bx\in\Dx, \params\in\Dparams) \mapsto
\argmax_{\ccc\in\Cset} \network_\ccc(x, \params)$, where $\params$ are
the parameters to be learnt, and $\network_\ccc(\bx, \params)$ is the
score given to the (input, class) pair $(\bx, \ccc)$ by a function
$\network:\Dx\times\Dparams\rightarrow \Dc$. We take $\network$ to be
a neural network, represented by a computation graph $G=(\nodes,
\edges)$, which is a directed acyclic graph with a single root node,
and each node $\nod\in\nodes$ takes values in
$\Re^{\Dout\nodind{\nod}}$ and is a function of its children in the graph,
  with learnable parameters $\params\nodind{\nod}$:
\begin{equation}
\label{eq:netnet}
\nod:\bx\mapsto\layerfunc\nodind{\nod}\big(\params\nodind{n},
\big(\nod'(\bx)\big)_{\nod':(\nod, \nod')\in\edges} \big)\,.
\end{equation}
The function $\network$ we want to learn is the root of $\graph$. The
training data $((\bx_i, \by_i))_{i=1}^m\in({\calX}\times{\cal Y})^m$
is an i.i.d. sample of ${\cal D}$, and we assume ${\cal
  X}\subset\Dx$ is compact. A function $\loss:\Dc\times\Cset
\rightarrow \Re$ measures the loss of $\network$ on an example
$(\bx,\by)$; in a single-label classification setting for instance, a
common choice for $\loss$ is the log-loss:
\begin{equation}
\label{eq:softmax}
\logsoftmax\big(\network(\bx, \params), \by\big) = - \network_{\by}(x,
\params) + \log \big(\sum_{\ccc\in\Cset} e^{\network_{\ccc}(x,
    \params)}\big).\!\!
\end{equation}
The arguments that we develop below depend only on the Lipschitz
constant of the loss, with respect to the norm of interest. Formally,
we assume that given a $p$-norm of interest $\normp{.}$, there is a
constant $\liplossp$ such that
\begin{equation}
\forall \bdc, \bdc' \in \Dc, \forall \ccc\in\Cset,
|\ell(\bdc, \ccc) - \ell(\bdc', \ccc)| \leq \liplossp\normp{\bdc-\bdc'}\,.
\end{equation}
For the log-loss of \eqref{eq:softmax}, we have $\liplosst \leq
\sqrt{2}$ and $\liplossinf \leq 2$. In the next subsection, we define
adversarial examples and the generalization performance of the
classifier. Then, we make the relationship between robustness to
adversarial examples and the lipschitz constant of the networks.

\subsection{Adversarial examples}
\label{sec:advex}
Given an input (train or test) example $(\bx, \by)$, an adversarial
example is a perturbation of the input pattern $\adv{\bx} = \bx +
\perturb{\bx}$ where $\perturb{\bx}$ is small enough so that
$\adv{\bx}$ is nearly undistinguishable from $\bx$ (at least from the
point of view of a human annotator), but has the network predict an
incorrect label. Given the network parameters and structure $g(.,
\params)$ and a $p$-norm, the adversarial example is formally defined as
\begin{equation}
\label{eq:adv}
\bxprime = \argmax_{\bxprime:\normp{\bxprime-\bx}\leq \adveps}
\loss\big(\network(\bxprime, \params), \by\big)\,,
\end{equation}
where $\adveps$ represents the strength of the adversary. Since the
optimization problem above is non-convex,
\citet{shaham2015understanding} propose to take the first
order taylor expansion of $\bx\mapsto\loss(\network(\bx, \params),
\by)$ to compute $\perturb{\bx}$ by solving
\begin{equation}
\label{eq:fgsm}
\bxprime = \argmax_{\bxprime:\normp{\bxprime-\bx}\leq \adveps}
\big(\nabla_{\bx} \loss(\network(\bx, \params),
\by)\big)^T (\bxprime-\bx)\,.
\end{equation}
If $p=\infty$, then $\bxprime = \bx + \adveps \text{sign}(\nabla_{\bx}
\loss(\network(\bx, \params), \by))$. This is the \emph{fast gradient
  sign method}. For the case $p=2$, we obtain $\bxprime = \bx + \adveps
\nabla_{\bx} \loss(\network(\bx, \params), \by)$. A more involved
method is the \emph{iterative fast gradient sign method}, in which
several gradient steps of \eqref{eq:fgsm} are performed with a smaller
stepsize to obtain a local minimum of \eqref{eq:adv}.

\subsection{Generalization with adversarial examples}

In the context of adversarial examples, there are two different
generalization errors of interest:
\begin{align*}
\Loss{W} &= \expect{(x,y)\sim{\cal D}}{\loss(\network(\bx, \params),
  \by)}\,,\\
\Lossadv{W}{p}{\adveps} &= \expect{(x,y)\sim{\cal
    D}}{\max_{\bxprime:\normp{\bxprime-\bx}\leq \adveps}
  \loss(\network(\bxprime, \params), \by)}\,.
\end{align*}
By definition, $\Loss{\bW}\leq
\Lossadv{\bW}{p}{\adveps}$ for every $p$ and
$\adveps\!>\!0$. Reciprocally, denoting by $\liplossp$ and $\lipgp$ the
Lipschitz constant (with respect to $\normp{.}$) of $\ell$ and $\network$
respectively, we have:
\begin{equation*}
\begin{split}
&\Lossadv{W}{p}{\adveps} \leq \Loss{W} \\
& + \expect{(x,y)\sim{\cal
    D}}{\max_{\bxprime:\normp{\bxprime-\bx}\leq
    \adveps}|\loss(\network(\bxprime, \params), \by) -
  \loss(\network(\bx, \params), \by)|}\\
&\phantom{\Lossadv{W}{p}{\adveps}}\leq \Loss{W}
+ \liplossp\lipgp\adveps\,.
\end{split}
\end{equation*}
This suggests that the sensitivity to
adversarial examples can be controlled by the Lipschitz constant of
the network. In the robustness framework of \cite{xu2012robustness},
the Lipschitz constant also controls the
difference between the average loss on the training set and the
generalization performance. More precisely, let us denote by
$\covnump{\coveps}$ the covering number of $\calX$ using
$\coveps$-balls for $\normp{.}$. Using $M = \sup_{\bx,\params, \by}
\loss(\network(\bx, \params), \by)$, Theorem 3 of
\cite{xu2012robustness} implies that for every $\delta\in(0, 1)$, with
probability $1-\delta$ over the i.i.d. sample
$((\bx_i,\by_i)_{i=1}^m$, we have:
\begin{align*}
\Loss{W} \leq & \frac{1}{m} \sum_{i=1}^m \loss(\network(\bx_i, \params), \by_i)\\
& + \liplossp\lipgp\coveps +  M\sqrt{\frac{2\nC\covnump{\frac{\coveps}{2}}\ln(2) - 2\ln(\delta)}{m}}\,.
\end{align*}
Since covering numbers of a $p$-norm ball in $\Dx$ increases
exponentially with $\Dx$, the bound above suggests that it is critical
to control the Lipschitz constant of $\network$, for both good
generalization and robustness to adversarial examples.

\subsection{Lipschitz constant of neural networks}

From the network structure we consider \eqref{eq:netnet}, for every
node $\nod\in\nodes$, we have (see below for the definition of
$\lipp\nodind{\nod, \nod'}$):
\begin{equation*}
\normp{\nod(\bx)-\nod(\bxprime)} \leq
\sum_{\nod':(\nod, \nod')\in\edges}
\lipp\nodind{\nod, \nod'}\normp{\nod'(\bx)-\nod'(\bxprime) }\,,
\end{equation*}
for any $\lipp\nodind{\nod, \nod'}$ that is greater than the worst
case variation of $\nod$ with respect to a change in its input
$\nod'(\bx)$. In particular we can take for $\lipp\nodind{\nod,
  \nod'}$ any value greater than the supremum over $\bx_0 \in \calX$
of the Lipschitz constant for $\norm{.}_p$ of the function
($\indic{\nod''=\nod'}$ is $1$ if $\nod''=\nod'$ and $0$ otherwise):
\begin{equation*}
\bx \mapsto
\layerfunc\nodind{\nod}\big(\params\nodind{n},
\big(\nod''(\bx_0 + \indic{\nod''=\nod'}(\bx-\bx_0))\big)
_{\nod'':(\nod, \nod'')\in\edges} \big)\,.
\end{equation*}
The Lipschitz constant of $\nod$, denoted by $\lipp\nodind{\nod}$
satisfies:
\begin{equation}
\label{eq:lipall}
\lipp\nodind{\nod} \leq \sum_{\nod':(\nod, \nod')\in\edges}
\lipp\nodind{\nod, \nod'}\lipp\nodind{\nod'}
\end{equation}
Thus, the Lipschitz constant of the network $g$ can grow exponentially
with its depth. We now give the Lipschitz constants of standard
layers as a function of their parameters:
\paragraph{Linear layers:} For layer
$\nod(x) = \params\nodind{\nod}\nod'(x)$ where $\nod'$ is the unique
child of $\nod$ in the graph, the Lipschitz constant for $\normp{.}$
is, by definition, the matrix norm of $\params\nodind{\nod}$ induced
by $\normp{.}$, which is usually denoted
$\normp{\params\nodind{\nod}}$ and defined by
\begin{equation*}
\normp{\params\nodind{\nod}} = \sup_{\bdc:\normp{\bdc}=1}
\normp{\params\nodind{\nod}\bdc}\,.
\end{equation*}
Then $\lipgt\nodind{\nod} =
\normt{\params\nodind{\nod}}\lipgt\nodind{\nod'}$, where
$\normt{\params\nodind{\nod}}$, called the spectral norm of
$\params\nodind{\nod}$, is the maximum singular value of
$\params\nodind{\nod}$. We also have $\lipginf\nodind{\nod} =
\norminf{\params\nodind{\nod}}\lipginf\nodind{\nod'}$, where
$\norminf{\params\nodind{\nod}} = \max_{i} \sum_j
|\params\nodind{\nod}_{ij}|$ is the maximum $1$-norm of the rows.
$\params\nodind{\nod}$.
\paragraph{Convolutional layers:}
To simplify notation, let us consider convolutions on 1D inputs
without striding, and we take the width of the convolution to be
$2\wid+1$ for $\wid\in\Nats$. To write convolutional layers in the same way
as linear layers, we first define an unfolding operator $\uU$, which
prepares the input $\bdc$, denoted by $\uU{\bdc}$. If the input has
length $T$ with $\Din$ inputs channels, the unfolding operator maps
$\bdc$ For a convolution of the unfolding of $\bdc$ considered as a
$T\times (2\wid+1)\Din$ matrix, its $j$-th column is:
\begin{equation}
\label{eq:unfold}
\uU_j(\bdc) = [\bdc_{j-\wid} ;  ... ; \bdc_{j+\wid}]\,,
\end{equation}
where ``;'' is the concatenation along the vertical axis (each
$\bdc_i$ is seen as a column $\Din$-dimensional vector), and $\bdc_i =
0$ if $i$ is out of bounds ($0$-padding).  A convolutional layer with
$\Dout$ output channels is then defined as
\begin{equation}
\nod(\bx) = \bW\nodind{\nod}\conv\nod'(\bx) = \bW\nodind{\nod}\uU(\nod'(\bx))\,,
\end{equation}
where $\bW\nodind{\nod}$ is a $\Dout\times(2\wid+1)\Din$ matrix. We
thus have $\lipgt\nodind{\nod} \leq
\normt{\bW}\normt{\uU(\nod'(\bx))}$. Since $\uU$ is a linear operator
that essentially repeats its input $(2\wid+1)$ times, we have
$\normt{\uU(\nod'(\bx)) - \uU(\nod'(\bxprime))}^2 \leq
(2\wid+1)\normt{\nod'(\bx) - \nod'(\bxprime)}^2$, so that $\lipgt\nodind{\nod} \leq
\sqrt{2\wid+1}\normt{\bW} \lipgt\nodind{\nod'}$.
Also, $\norminf{\uU(\nod'(\bx)) - \uU(\nod'(\bxprime))} =
\norminf{\nod'(\bx) - \nod'(\bxprime)}$, and so for a convolutional
layer, $\lipginf\nodind{\nod} \leq
\norminf{\params\nodind{\nod}}\lipginf\nodind{\nod'}$.

\paragraph{Aggregation layers/transfer functions:}
Layers that perform the sum of their inputs, as in Residual Netowrks
\cite{he2016deep}, fall in the case where the
values $\lipp\nodind{\nod, \nod'}$ in \eqref{eq:lipall} come into
play. For a node $\nod$ that sums its inputs, we have $\lipp\nodind{\nod, \nod'}=1$,
and thus $\lipp\nodind{\nod} \leq \sum_{\nod':(\nod,
  \nod')\in\edges}\lipp\nodind{\nod'}$ .
If $\nod$ is a tranfer function layer (e.g., an element-wise
application of ReLU) we can check that $\lipp\nodind{\nod} \leq
\lipp\nodind{\nod'}$, where $\nod'$ is the input node, as soon as the
Lipschitz constant of the transfer function (as a function
$\Re\rightarrow\Re$) is $\leq 1$.

\section{Parseval networks}

Parseval regularization, which we introduce in this section, is a
regularization scheme to make deep neural networks robust, by
constraining the Lipschitz constant \eqref{eq:lipall} of each hidden
layer to be smaller than one, assuming the Lipschitz constant of
children nodes is smaller than one. That way, we avoid the exponential
growth of the Lipschitz constant, and a usual regularization scheme
(i.e., weight decay) at the last layer then controls the overall
Lipschitz constant of the network. To enforce these constraints in
practice, Parseval networks use two ideas: maintaining orthonormal
rows in linear/convolutional layers, and performing convex
combinations in aggregation layers. Below, we first explain the
rationale of these constraints and then describe our approach to
efficiently enforce the constraints during training.

\subsection{Parseval Regularization}
\paragraph{Orthonormality of weight matrices:}
For linear layers, we need to maintain the spectral norm of the weight
matrix at $1$. Computing the largest singular value of weight matrices
is not practical in an SGD setting unless the rows of the matrix are
kept orthogonal. For a weight matrix
$\bW\in\Re^{\Dout\times\Din}$ with $\Dout\leq\Din$, Parseval
regularization maintains $\bW^T \bW \approx I_{\Dout\times\Dout}$,
where $I$ refers to the identity matrix. $\bW$ is then approximately
a \emph{Parseval tight frame}~\cite{kovavcevic2008introduction}, hence
the name of Parseval networks.  For convolutional layers, the matrix
$\bW\in\Re^{\Dout\times(2\wid+1)\Din}$ is constrained to be a Parseval
tight frame (with the notations of the previous section), and the
output is rescaled by a factor $(2\wid+1)^{-1/2}$. This maintains all
singular values of $\bW$ to $(2\wid+1)^{-1/2}$, so that
$\lipgt\nodind{\nod} \leq \lipgt\nodind{\nod'}$ where $\nod'$ is the
input node. More generally, keeping the rows of weight matrices orthogonal makes
it possible to control both the spectral norm and the $\norminf{.}$ of
a weight matrix through the norm of its individual rows. Robustness
for $\norminf{.}$ is achieved by rescaling the rows so that their
$1$-norm is smaller than $1$. For now, we only experimented with
constraints on the $2$-norm of the rows, so we aim for robustness in the sense of
$\normt{.}$.

\begin{remark}[Orthogonality is required]
Without orthogonality, constraints on the $2$-norm of the rows of
weight matrices are not sufficient to control the spectral
norm. Parseval networks are thus fundamentally different from weight
normalization \cite{salimans2016weight}.
\end{remark}

\paragraph{Aggregation Layers:}
In parseval networks, aggregation layers do not make the sum of
their inputs, but rather take a convex combination of them:
 \begin{equation}
 \label{eq:cvx}
 \nod(\bx) = \sum_{\nod':(\nod, \nod')\in\edges}
\alpha\nodind{\nod, \nod'} \nod'(\bx)
\end{equation}
with $\sum_{\nod':(\nod,
  \nod')\in\edges} \alpha\nodind{\nod, \nod'} = 1$ and
$\alpha\nodind{\nod, \nod'} \geq 0$. The parameters $\alpha\nodind{\nod, \nod'}$ are learnt,
but using \eqref{eq:lipall}, these constraint guarantee that
$\lipgp\nodind{\nod} \leq 1$ as soon as the children satisfy the inequality
for the same $p$-norm.

\subsection{Parseval Training}

\paragraph{Orthonormality constraints:}The first significant difference between Parseval networks and its vanilla counterpart is the orthogonality constraint on the weight matrices.
This requirement calls for an optimization algorithm on the manifold of orthogonal matrices, namely the \emph{Stiefel manifold}.
Optimization on matrix manifolds is a well-studied topic (see~\cite{absil2009optimization} for a comprehensive survey).
The simplest first-order geometry approaches consist in optimizing the unconstrained function of interest by moving in the direction of steepest descent (given by the gradient of the function) while at the same time staying on the manifold.
To guarantee that we remain in the manifold after every parameter update, we need to define a retraction operator.
There exist several pullback operators for embedded submanifolds such as the \emph{Stiefel manifold} based for example on Cayley transforms~\cite{absil2009optimization}.
However, when learning the parameters of neural networks, these methods are computationally prohibitive. To overcome this difficulty, we use an approximate operator derived from the following layer-wise regularizer of weight matrices to ensure their \emph{parseval tightness}~\cite{kovavcevic2008introduction}:
\begin{equation}
    R_\beta(W_k) = \frac{\beta}{2}\|W_k^\top W_k - I\|_2^2.
\end{equation}

\begin{algorithm}[t]
\small
    $\Theta = \{W_k,\boldsymbol{\alpha}_k\}_{k=1}^K$, $e \leftarrow 0$\\
    \While{ $e \leq E$ }
    {
        Sample a minibatch $\{(x_i, y_i)\}_{i=1}^B$. \\
        \For{$k \in \{1, \dots, K\}$}
        {
            Compute the gradient: $G_{W_k}\leftarrow \nabla_{W_k} \ell(\Theta,\{(x_i, y_i)\})$, $G_{\boldsymbol{\alpha}_k} \leftarrow \nabla_{\boldsymbol{\alpha}_k} \ell(\Theta,\{(x_i, y_i)\})$.\\
            Update the parameters:\\
             $W_k \leftarrow W_k - \epsilon\cdot G_{W_k}$ \\
            $\boldsymbol{\alpha}_k \leftarrow \boldsymbol{\alpha}_k - \epsilon\cdot G_{\boldsymbol{\alpha}_k}$. \\
            \If{hidden layer}{
                Sample a subset $S$ of rows of $W_k$. \\
                Projection: \\
                $W_S \leftarrow (1 + \beta) W_S - \beta W_S W_S^\top W_S$. \\
                $\boldsymbol{\alpha}_k \leftarrow  \argmin_{\boldsymbol{\gamma}\in \Delta^{K-1}} \norm{\boldsymbol{\alpha}_K - \boldsymbol{\gamma}}_2^2$
            }
        }
        $e \leftarrow e+1$.
    }
    \caption{
      	Parseval Training
    }
    \label{alg:parseval}
\end{algorithm}

Optimizing $R_\beta(W_k)$ to convergence after every gradient descent step (w.r.t the main objective) guarantees us to stay on the desired manifold but this is an expensive procedure. 
Moreover, it may result in parameters that are far from the ones obtained after the main gradient update.
We use two approximations to make the algorithm more efficient:
First, we only do one step of descent on the function $R_\alpha(W_k)$.
The gradient of this regularization term
is $\nabla_{W_k} R_\beta(W_k) = \beta (W_k W_k^\top - I) W_k$.
Consequently, after every main update we perform the following secondary update:
\begin{equation}
    \label{eq:retract}
    W_k \leftarrow (1+\beta)  W_k - \beta W_k W_k^\top W_k.
\end{equation}
Optionally, instead of updating the whole matrix, one can randomly select a subset $S$ of rows and perform the update from Eq.~(\ref{eq:retract}) on the submatrix composed of rows indexed by $S$. This sampling based approach reduces the overall complexity to $\mathcal{O}(|S|^2 d)$. Provided the rows are carefully sampled, the procedure is an accurate Monte Carlo approximation of the regularizer loss function~\cite{drineas2006fast}. The optimal sampling probabilities, also called statistical leverages are approximately equal if we start from an orthogonal matrix and (approximately) stay on the manifold throughout the optimization since they are proportional to the eigenvalues of $W$~\cite{mahoney2011randomized}.  Therefore, we can sample a subset of columns uniformly at random when applying this projection step.

While the full update does not result in an increased overhead for convolutional layers, the picture can be very different for large fully connected layers making the sampling approach computationally more appealing for such layers. We show in the experiments that the weight matrices resulting from this procedure are (quasi)-orthogonal. Also, note that quasi-orthogonalization procedures similar to the one described here have been successfully used previously in the context of learning overcomplete representations with independent component analysis~\cite{hyvarinen2000independent}.

\paragraph{Convexity constraints in aggregation layers:} In Parseval networks, aggregation layers output a \emph{convex combination} of their inputs instead of e.g., their sum as in Residual networks~\cite{he2016deep}. 
For an aggregation node $n$ of the network, let us denote by
$\boldsymbol{\alpha} = (\alpha^{(\nod, \nod')})_{\nod': (\nod,
  \nod')\in\edges}$ the $K$-size vector of coefficients used for the
convex combination output by the layer.
To ensure that the Lipschitz constant at the node $\nod$
is such that $\lipgp\nodind{\nod} \leq 1$, the constraints
of~\ref{eq:cvx} call for a euclidean projection of
$\boldsymbol{\alpha}$ onto the positive simplex after a gradient update:
\begin{equation}
\boldsymbol{\alpha}^* =  \argmin_{\boldsymbol{\gamma}\in \Delta^{K-1}} \norm{\boldsymbol{\alpha} - \boldsymbol{\gamma}}_2^2\,,
\end{equation}
where $\Delta^{K-1} = \{\boldsymbol{\gamma} \in \mathbb{R}^K| \boldsymbol{1}^\top\boldsymbol{\gamma}=1, \boldsymbol{\gamma} \ge 0\}$. This is a well studied problem~\cite{michelot1986finite,pardalos1990algorithm,duchi2008efficient,condat2016fast}. Its solution is of the form: $\alpha^*_i = \max(0, \alpha_i - \tau(\boldsymbol{\alpha}))$, with $\tau: \mathbb{R}^K \rightarrow \mathbb{R}$ the unique function satisfying $\sum_i(x_i - \tau(\boldsymbol{\alpha})) = 1$ for every $\boldsymbol{x}\in \mathbb{R}^K$. Therefore, the solution essentially boils down to a \emph{soft thresholding} operation. If we denote  $\alpha_1 \ge \alpha_{2} \ge \dots \alpha_K$ the sorted coefficients and $k(\boldsymbol{\alpha}) = \max\{k \in (1,\dots,K)| 1+k\alpha_k > \sum_{j \le k}\alpha_j\}$, the optimal thresholding is given by~\cite{duchi2008efficient}:
\begin{equation}
\tau(\boldsymbol{\alpha}) = \frac{(\sum_{j \le k(\boldsymbol{\alpha})}\alpha_j) - 1} {k(\boldsymbol{\alpha})}
\end{equation}
Consequently, the complexity of the projection is $O(K\log(K))$ since it is only dominated by the sorting of the coefficients and is typically cheap because aggregation nodes will only have few children in practice (e.g. 2). If the number of children is large, there exist efficient linear time algorithms for finding the optimal thresholding $\tau(\boldsymbol{\alpha})$~\cite{michelot1986finite,pardalos1990algorithm,condat2016fast}. In this work,  we use the method detailed above~\cite{duchi2008efficient} to perform the projection of the coefficient $\boldsymbol{\alpha}$ after every gradient update step.


\section{Experimental evaluation}

We evaluate the effectiveness of Parseval networks on well-established image classification benchmark datasets namely MNIST, CIFAR-10, CIFAR-100~\cite{krizhevsky2009learning} and Street View House Numbers (SVHN)~\cite{netzer2011reading}. We train both fully connected networks and wide residual networks. The details of the datasets, the models, and the training routines are summarized below.

\begin{figure}[t]
    \centering
    \includegraphics[width=0.115\textwidth]{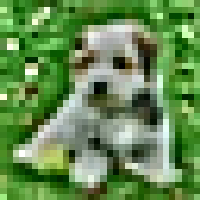}
    \includegraphics[width=0.115\textwidth]{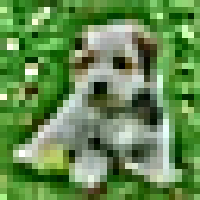}
    \includegraphics[width=0.115\textwidth]{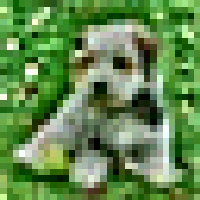}
    \includegraphics[width=0.115\textwidth]{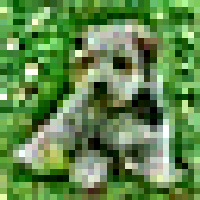} \\
    \vspace{0.4em}
    \includegraphics[width=0.115\textwidth]{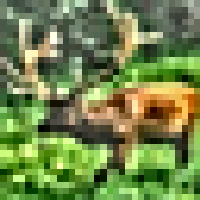}
    \includegraphics[width=0.115\textwidth]{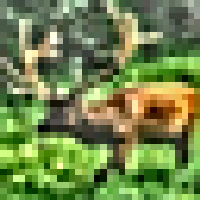}
    \includegraphics[width=0.115\textwidth]{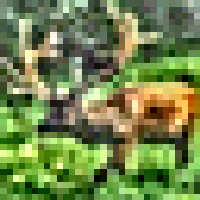}
    \includegraphics[width=0.115\textwidth]{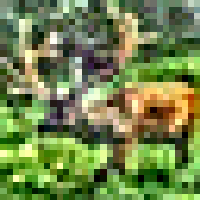}
    \vspace{0.4em}
   \includegraphics[width=0.115\textwidth]{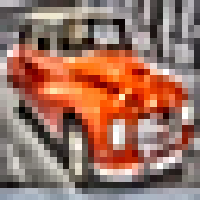}
   \includegraphics[width=0.115\textwidth]{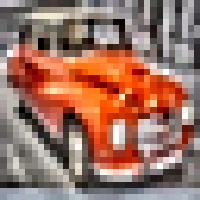}
   \includegraphics[width=0.115\textwidth]{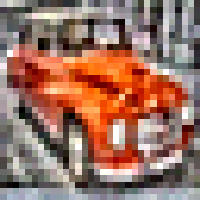}
   \includegraphics[width=0.115\textwidth]{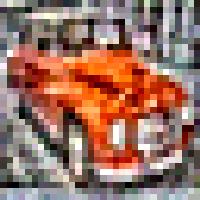} \\

    \caption{
        Sample images from the CIFAR-10 dataset, with corresponding adversarial examples.
        We show the original image and adversarial versions for SNR values of 24.7, 12.1 and 7.8.
    }
    \label{fig:images}
    \vspace{-1em}
\end{figure}

\subsection{Datasets}

\myparagraph{CIFAR.}
 Each of the CIFAR datasets is composed of 60K natural scene color images of size $32\times32$ split between 50K training images and 10K test images. CIFAR-10 and CIFAR-100 have respectively 10 and 100 classes. For these two datasets, we adopt the following standard preprocessing and data augmentation scheme~\cite{lin2013network,he2016deep,huang2016densely,zagoruyko2016wide}: Each training image is first zero-padded with 4 pixels on each side. The resulting image is randomly cropped to produce a new $32\times32$ image which is subsequently horizontally flipped with probability $0.5$. We also normalize every image with the mean and standard deviation of its channels. Following the same practice as~\cite{huang2016densely}, we initially use 5K images from the training as a validation set. Next, we train de novo the best model on the full set of 50K images and report the results on the test set.
\myparagraph{SVHN} The Street View House Number dataset is a set of $32\times32$ color digit images officially split into 73257 training images and 26032 test images. Following common practice~\cite{zagoruyko2016wide,he2016deep,huang2016densely,huang2016deep}, we randomly sample 10000 images from the available extra set of about $600K$ images as a validation set and combine the rest of the pictures with the official training set for a total number of 594388 training images. We divide the pixel values by 255 as a preprocessing step and report the test set performance of the best performing model on the validation set.
\subsection{Models and Implementation details}

\subsubsection{ConvNets}
\myparagraph{Models.}
For the CIFAR and SVHN datasets, we trained wide residual networks~\cite{zagoruyko2016wide} as they perform on par with standard resnets~\cite{he2016deep} while being faster to train thanks to a reduced depth. We used wide resnets of depth 28 and width 10 for both CIFAR-10 and CIFAR-100. For SVHN we used wide resnet of depth 16 and width 4. For each architecture, we compare Parseval networks with the vanilla model trained with standard regularization both in the adversarial and the non-adversarial training settings.

\myparagraph{Training.}
We train the networks with stochastic gradient descent using a momentum of 0.9. On CIFAR datasets, the initial learning rate is set to $0.1$ and scaled by a factor of $0.2$ after epochs 60, 120 and 160, for a total number of 200 epochs. We used mini-batches of size 128. For SVHN, we trained the models with mini-batches of size 128 for 160 epochs starting with a learning rate of 0.01 and decreasing it by a factor of $0.1$ at epochs 80 and 120.
For all the vanilla models, we applied by default weight decay regularization (with parameter $\lambda=0.0005$) together with batch normalization and dropout since this combination resulted in better accuracy and increased robustness in preliminary experiments. The dropout rate use is 0.3 for CIFAR and 0.4 for SVHN. For Parseval regularized models, we choose the value of the retraction parameter to be $\beta=0.0003$ for CIFAR datasets and $\beta=0.0001$ for SVHN based on the performance on the validation set. In all cases, We also adversarially trained each of the models on CIFAR-10 and CIFAR-100 following the guidelines in~\cite{goodfellow2015explaining,shaham2015understanding,kurakin2016adversarial}. In particular, we replace $50\%$ of the examples of every minibatch by their adversarially perturbed version generated using the one-step method to avoid label leaking~\cite{kurakin2016adversarial}. For each mini-batch, the magnitude of the adversarial perturbation is obtained by sampling from a truncated Gaussian centered at $0$ with standard deviation 2.

\subsubsection{Fully Connected}
\myparagraph{Model.}
We also train feedforward networks composed of 4 fully connected hidden layers of size 2048 and a classification layer.
The input to these networks are images unrolled into a $C \times 1024$ dimensional vector where $C$ is the number of channels.
We used these models on MNIST and CIFAR-10 mainly to demonstrate that the proposed approach is also useful on non-convolutional networks.
We compare a Parseval networks to vanilla models with and without weight decay regularization. For adversarially trained models, we follow the guidelines previously described for the convolutional networks.

\myparagraph{Training.}
We train the models with SGD and divide the learning rate by two every 10 epochs.
We use mini-batches of size 100 and train the model for 50 epochs.
We chose the hyperparameters on the validation set and re-train the model on the union of the training and validation sets.
The hyperparameters are $\beta$, the size of the row subset $S$,
 the learning rate and its decrease rate.
Using a subset $S$ of $30\%$ of all the rows of each of weight matrix for the retraction step worked well in practice.

\subsection{Results}
\subsubsection{(Quasi)-orthogonality.}
We first validate that Parseval training
(Algorithm~\ref{alg:parseval}) indeed yields (near)-orthonormal weight
matrices.
To do so, we analyze
the spectrum of the weight matrices of the different models by
plotting the histograms of their singular values, and compare these histograms for Parseval networks to networks trained using standard SGD with and without weight decay (\textbf{SGD-wd} and \textbf{SGD}).

\begin{figure}[t]
    \centering
    \includegraphics[width=0.49\linewidth]{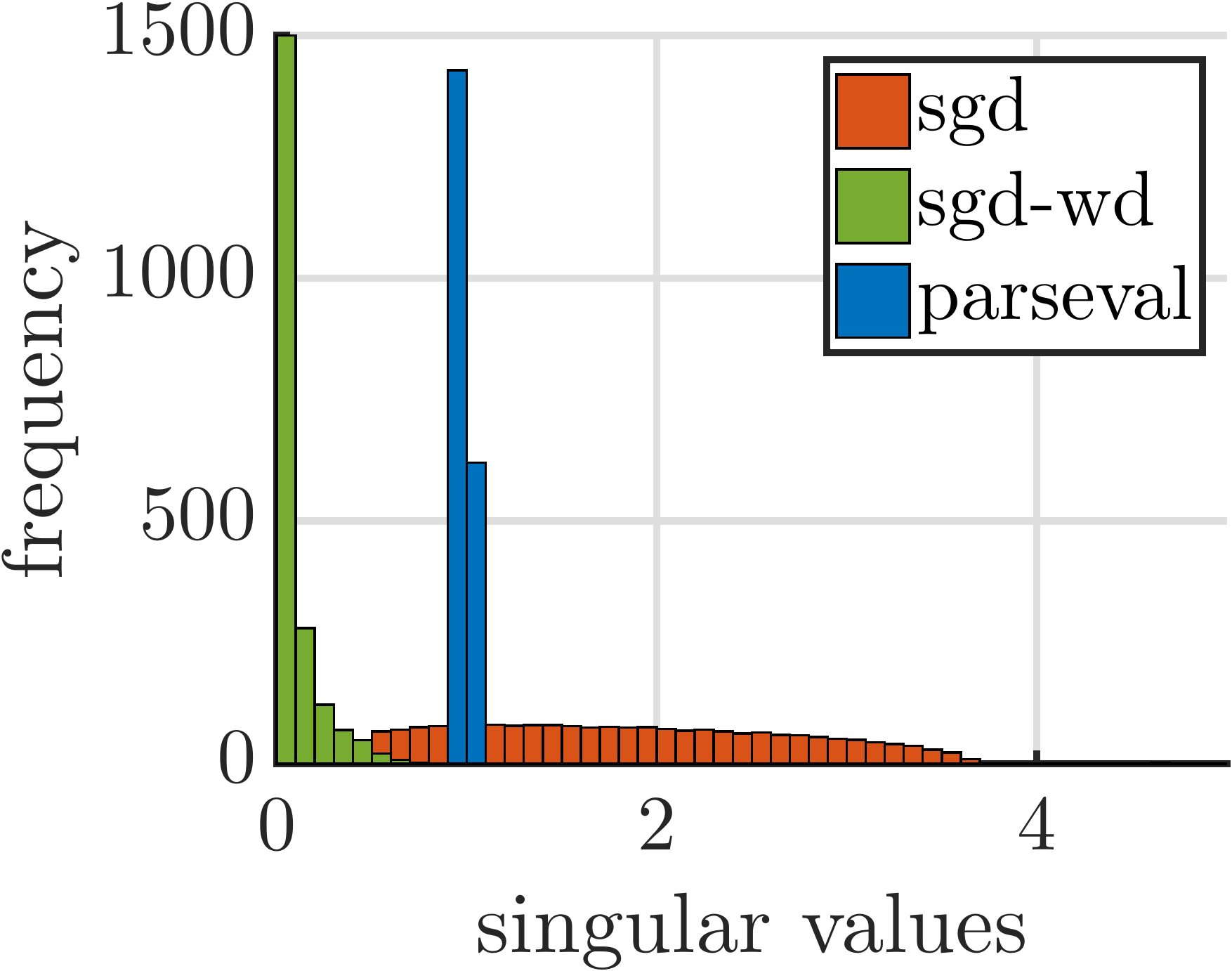}
    \includegraphics[width=0.49\linewidth]{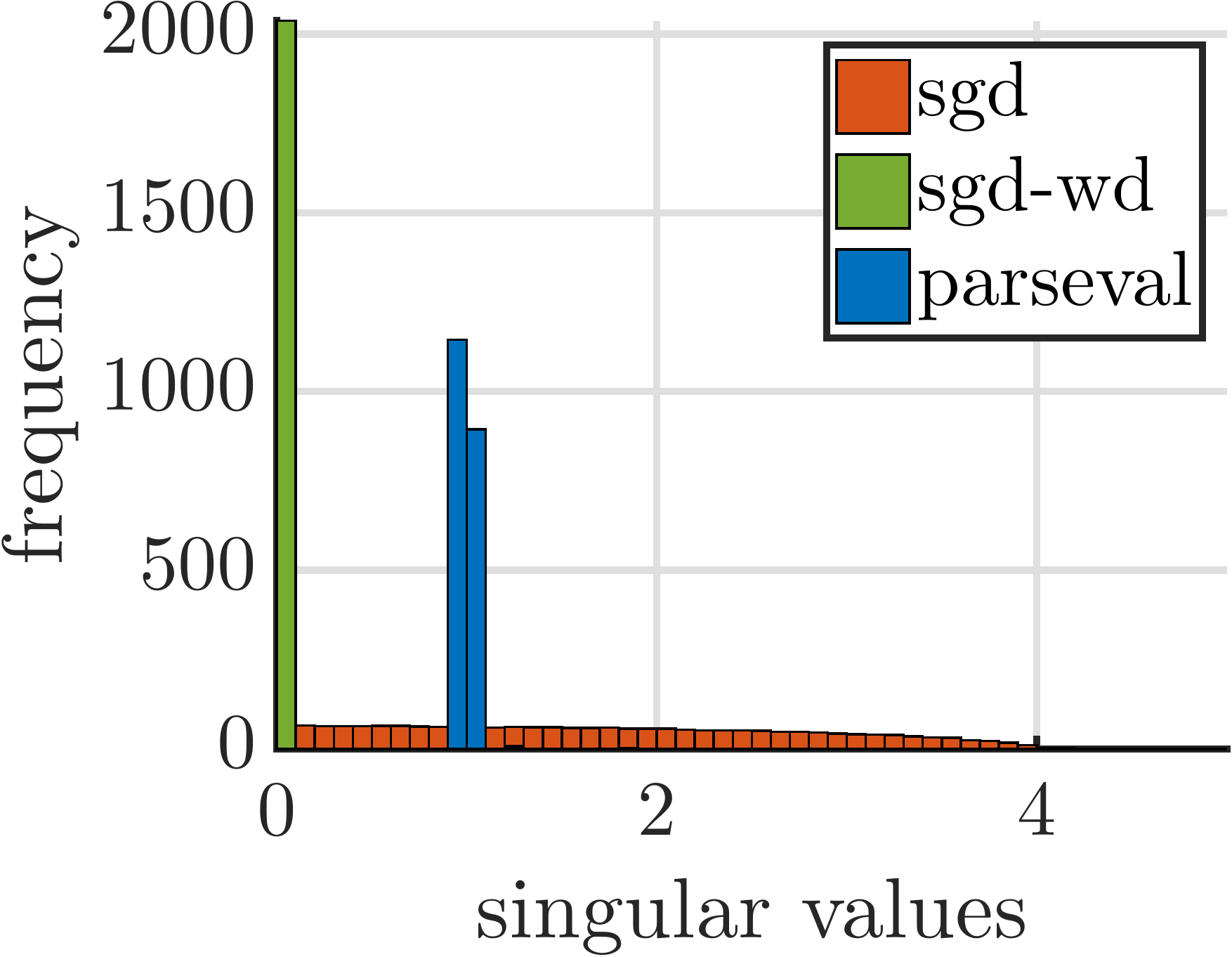}
    \caption{
        Histograms of the singular values of the weight matrices at layers 1 and 4 of our network in CIFAR-10.
    }
    \vspace{-1em}
    \label{fig:singular-values}
\end{figure}

The histograms representing the distribution of singular values at
layers 1 and 4 for the fully connected network (using $S=30\%$)
trained on the dataset CIFAR-10 are shown in
Fig.~\ref{fig:singular-values} (the figures for convolutional networks
are similar).  The singular values obtained with our method are
tightly concentrated around 1.  This experiment confirms that the weight
matrices produced by the proposed optimization procedure are (almost)
orthonormal.  The distribution of the singular values of the weight
matrices obtained with SGD has a lot more variance, with nearly as
many small values as large ones.  Adding weight decay to standard SGD
leads to a sparse spectrum for the weight matrices, especially in the
higher layers of the network suggesting a low-rank structure.  This observation
has motivated recent work on compressing deep neural
networks~\cite{denton2014exploiting}.

\subsubsection{Robustness to adversarial noise.}
We evaluate the robustness of the models to adversarial noise by
generating adversarial examples from the test set, for
various magnitudes of the noise vector.  Following common
practice~\cite{kurakin2016adversarial}, we use the fast gradient sign
method to generate the adversarial examples (using $\norminf{.}$, see
Section \ref{sec:advex}). Since these adversarial
examples transfer from one network to the other, the fast
gradient sign method allows to benchmark the network for reasonable
settings where the opponent does not know 
 the network.
We report the accuracy of each model as a function of the magnitude of the noise.
To make the results easier to interpret, we compute the corresponding Signal to Noise Ratio (SNR).
For an input $x$ and perturbation $\delta_x$, the SNR is defined as $\text{SNR}(x, \delta_x) = 20 \log_{10} \frac{\|x\|_2}{\|\delta_x\|_2}$.
We show some adversarial examples in~Fig.~\ref{fig:images}.
\begin{figure}[t]
    \centering
    \includegraphics[width=0.49\linewidth]{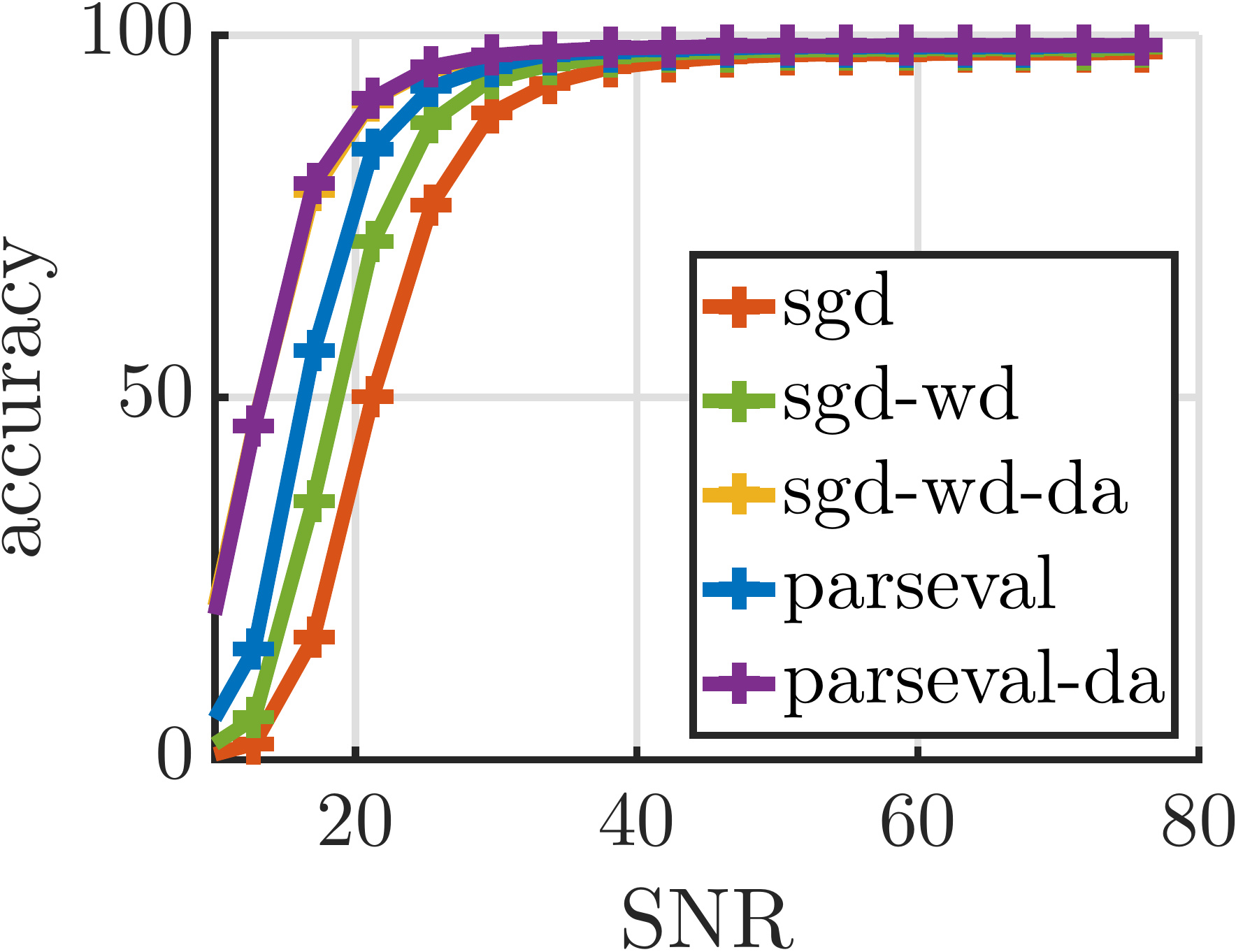}
    \includegraphics[width=0.49\linewidth]{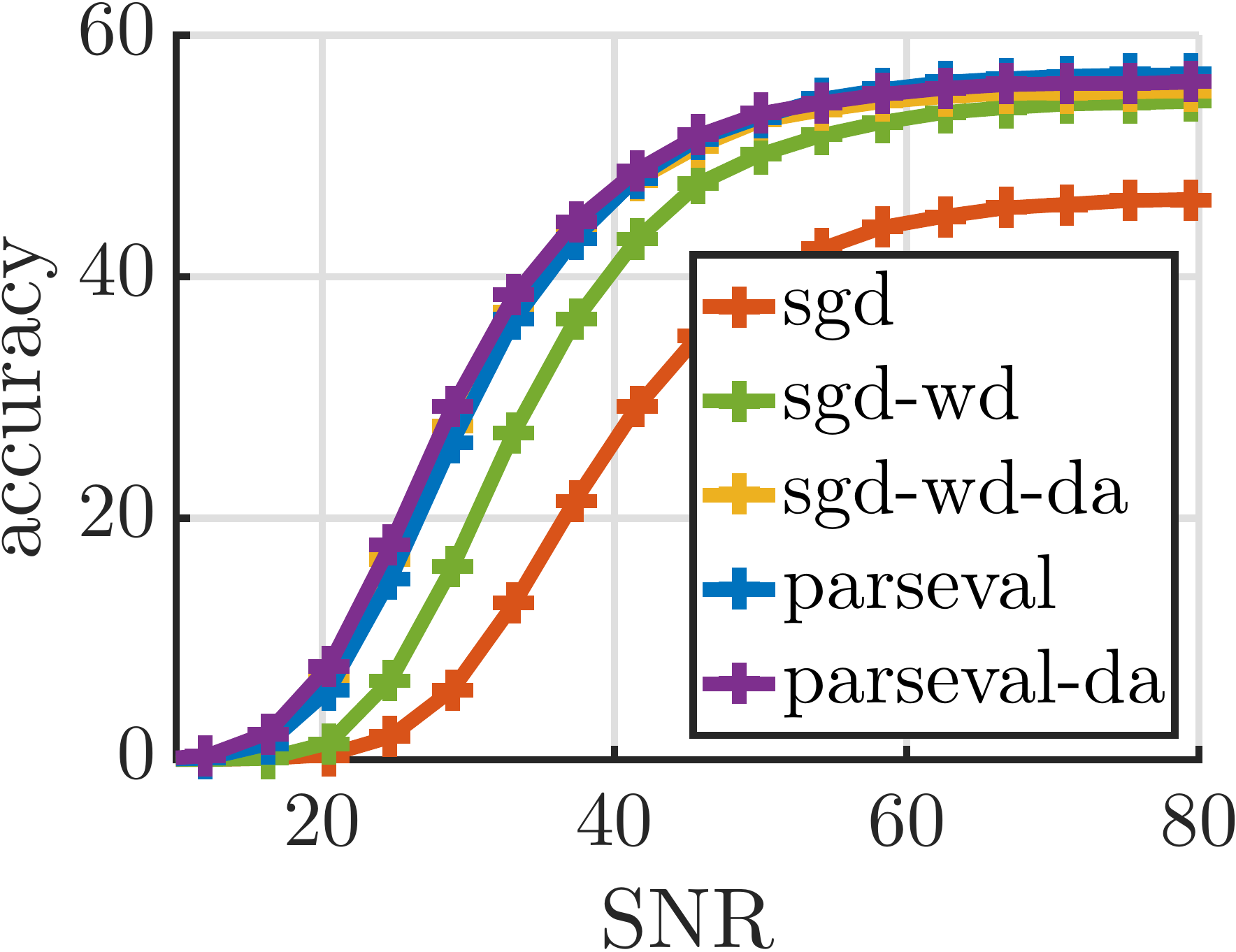}
    \caption{
        Performance of the models for various magnitudes of adversarial noise on MNIST (left) and CIFAR-10 (right).
            }
    \label{fig:all-noise}
   \vspace{-1em}
\end{figure}

\begin{table}[h!]
\begin{center}
\caption{
Classification accuracy of the models on CIFAR-10 and CIFAR-100 with the (combination of) various regularization scheme.
$\epsilon$ represents here the value of the signal to noise ratio (SNR). At $\epsilon=30$, an adversarially perturbed image is perceptible by a human. For each dataset, the top 3 rows report results for non-adversarial training and the bottom 3 rows report results for adversarial training. 
}\vspace{1.0ex}
\label{tab:big_table}
\bgroup\scriptsize
\def\arraystretch{1.4}
\begin{tabular}{|c|l|c|c|c|c|c|}
\hline
{\bf }
& {\bf Model}
& {\bf Clean} & {\bf $\epsilon\approx50$} & {\bf $\epsilon\approx45$} & {\bf $\epsilon\approx40$} & {\bf $\epsilon\approx33$} \\
\hline
\multirow{6}{*}{\rotatebox{90}{CIFAR-10}}
& Vanilla
& 95.63         & 90.16         & 85.97         & 76.62         & 67.21      \\
& Parseval(OC)
& 95.82         & 91.85         & 88.56         & 78.79         & 61.38      \\
& Parseval
& \textbf{96.28}          & \textbf{93.03}         & \textbf{90.40}         & \textbf{81.76}         & \textbf{69.10}      \\
\cline{2-7}
\cline{2-7}
& Vanilla
& 95.49         & 91.17         & 88.90         & 86.75         & 84.87      \\
& Parseval(OC)
& 95.59         & 92.31         & 90.00         & \textbf{87.02}         & \textbf{85.23}      \\
& Parseval
& \textbf{96.08}          & \textbf{92.51}         & \textbf{90.05}         & 86.89         & 84.53      \\
\hline\hline
\multirow{6}{*}{\rotatebox{90}{CIFAR-100}}
& Vanilla
& 79.70         & 65.76         & 57.27         & 44.62         & 34.49      \\
&Parseval(OC)
& 81.07         & 70.33         & 63.78         & 49.97         & 32.99      \\
& Parseval
& \textbf{80.72}          & \textbf{72.43}         & \textbf{66.41}         & \textbf{55.41}         & \textbf{41.19}      \\
\cline{2-7}
\cline{2-7}
& Vanilla
& 79.23         & 67.06         & 62.53         & 56.71         & 51.78      \\
& Parseval(OC)
& \textbf{80.34}         & 69.27         & 62.93         & 53.21         & \textbf{52.60}      \\
& Parseval
& 80.19          & \textbf{73.41}         & \textbf{67.16}         & \textbf{58.86}         & 39.56      \\
\hline\hline
\multirow{3}{*}{\rotatebox{90}{SVHN}}
& Vanilla
& \textbf{98.38}        & 97.04        & 95.18        & 92.71        & 88.11     \\
& Parseval(OC)
& 97.91       & \textbf{97.55}        & \textbf{96.35}         & \textbf{93.73}         & \textbf{89.09}      \\
& Parseval
& 98.13        & 97.86        & 96.19         & 93.55         & 88.47      \\
\cline{2-7}
\hline
\end{tabular}
\egroup
\end{center}
\vspace{-1ex}
\end{table}

\paragraph{Fully Connected Nets.} Figure~\ref{fig:all-noise} depicts a comparison of Parseval and vanilla networks with and without adversarial training at various noise levels.
On both MNIST and CIFAR-10, Parseval networks consistently outperforms weight decay regularization. In addition, it is as robust as adversarial training (SGD-wd-da) on
CIFAR-10. Combining Parseval Networks and adversarial training results in the most robust method on MNIST.

\paragraph{ResNets.}
Table~\ref{tab:big_table} summarizes the results of our experiments
with wide residual Parseval and vanilla networks on CIFAR-10,
CIFAR-100 and SVHN. In the table, we denote Parseval(OC) the Parseval
network with orthogonality constraint and without using a convex
combination in aggregation layers. Parseval indicates the configuration
where both of the orthogonality and convexity constraints are used. We
first observe that Parseval networks outperform vanilla ones on all
datasets on the clean examples and match the state of the art
performances on CIFAR-10 $(96.28\%)$ and SVHN $(98.44\%)$. On
CIFAR-100, when we use Parseval wide Resnet of depth 40 instead of 28,
we achieve an accuracy of $81.76\%$. In comparison, the best
performance achieved by a vanilla wide resnet~\cite{zagoruyko2016wide}
and a pre-activation resnet~\cite{he2016deep} are respectively
$81.12\%$ and $77.29\%$. Therefore, our proposal is a useful
regularizer for legitimate examples. Also note that in most cases, Parseval
networks combining both the orthogonality constraint and the convexity
constraint is superior to use the orthogonality constraint solely.

The results presented in the table validate our most important claim: Parseval
networks significantly improve the robustness of vanilla models to
adversarial examples. When no adversarial training is used, the gap in
accuracy between the two methods is significant (particularly in the
high noise scenario). For an SNR value of $40$, the best Parseval
network achieves $55.41\%$ accuracy while the best vanilla model is
at $44.62\%$. When the models are adversarially trained,
Parseval networks remain superior to vanilla models in most
cases. Interestingly, adversarial training only slightly improves the
robustness of Parseval networks in low noise setting (e.g. SNR values
of 45-50) and sometimes even deteriorates it (e.g. on CIFAR-10). In
contrast, combining adversarial training and Parseval networks is an
effective approach in the high noise setting. This result suggests that thanks
to the particular form of regularizer (controlling the Lipschitz constant
of the network), Parseval networks achieves robustness to adversarial
examples located in the immediate vicinity of each data point.
Therefore, adversarial training only helps for adversarial examples
found further away from the legitimate patterns. This observation
holds consistently across the datasets considered in this study.
\subsubsection{Better use of capacity}

\begin{table}[t]
\scriptsize
    \centering
    \caption{\small
        Number of dimensions (in $\%$ of the total dimension) necessary to capture 99\% of the covariance of the activations.
    }
    \vspace{0.5em}
    \begin{tabular}{r c rr c rr c rr c rr}
        \toprule
                && \multicolumn{2}{c}{SGD-wd} && \multicolumn{2}{c}{SGD-wd-da} && \multicolumn{2}{c}{Parseval} \\
                \cmidrule{3-4} \cmidrule{6-7} \cmidrule{9-10}
                && all & class && all & class && all & class \\
        \midrule
        Layer 1 &&  72.6 &   34.7 &&    73.6 &   34.7 &&   89.0 &   38.4 \\
        Layer 2 &&   1.5 &    1.3 &&    1.5 &    1.3 &&   82.6 &   38.2 \\
        Layer 3 &&   0.5 &    0.5 &&      0.4 &    0.4 &&   81.9 &   30.6 \\
        Layer 4 &&   0.5 &    0.4 &&    0.4 &    0.4 &&   56.0 &   19.3 \\
        \bottomrule
    \end{tabular}
    \label{tab:pca}
\end{table}
Given the distribution of singular values observed in Figure~\ref{fig:singular-values}, we want to analyze the intrinsic dimensionality of the representation learned by the different networks at every layer. To that end, we use the local covariance dimension~\cite{dasgupta2008random} which can be measured from the covariance matrix of the data. For each layer $k$ of the fully connected network, we compute the activation's empirical covariance matrix $\frac{1}{n} \sum_{i=1}^n \phi_k(x) \phi_k(x)^\top$ and obtain its sorted eigenvalues $\sigma_1 \geq \dots \geq \sigma_d$.
For each method and each layer, we select the smallest integer $p$ such that $\sum_{i=1}^p \sigma_i \geq 0.99 \sum_{i=1}^d \sigma_i$.
This gives us the number of dimensions that we need to explain $99\%$ of the covariance.
We can also compute the same quantity for the examples of each class, by only considering in the empirical estimation of the covariance of the examples $x_i$ such that $y_i = c$.
We report these numbers both on all examples and the per-class average on CIFAR-10 in Table~\ref{tab:pca}.

Table~\ref{tab:pca} shows that the local covariance dimension of all the data is consistently higher for Parseval networks than all the other approaches at any layer of the network.
SGD-wd-da contracts all the data in very low dimensional spaces at the upper levels of the network by using only $0.4\%$ of the total dimension (layer 3 and 4) while Parseval networks use about $81\%$ and  $56\%$ at of the whole dimension respectively in the same layers. This is intriguing given that SGD-wd-da also increases the robustness of the network, apparently not in the same way as Parseval networks. For the average local covariance dimension of the classes, SGD-wd-da contracts each class into the same dimensionality as it contracts all the data at the upper layers of the network. For Parseval, the data of each class is contracted in about $30\%$ and $19\%$ of the overall dimension. These results suggest that Parseval contracts the data of each class in a lower dimensional manifold (compared to the intrinsic dimensionality of the whole data) hence making classification easier.
\subsubsection{faster convergence}

\begin{figure}[t]
    \centering
    \includegraphics[width=0.49\linewidth]{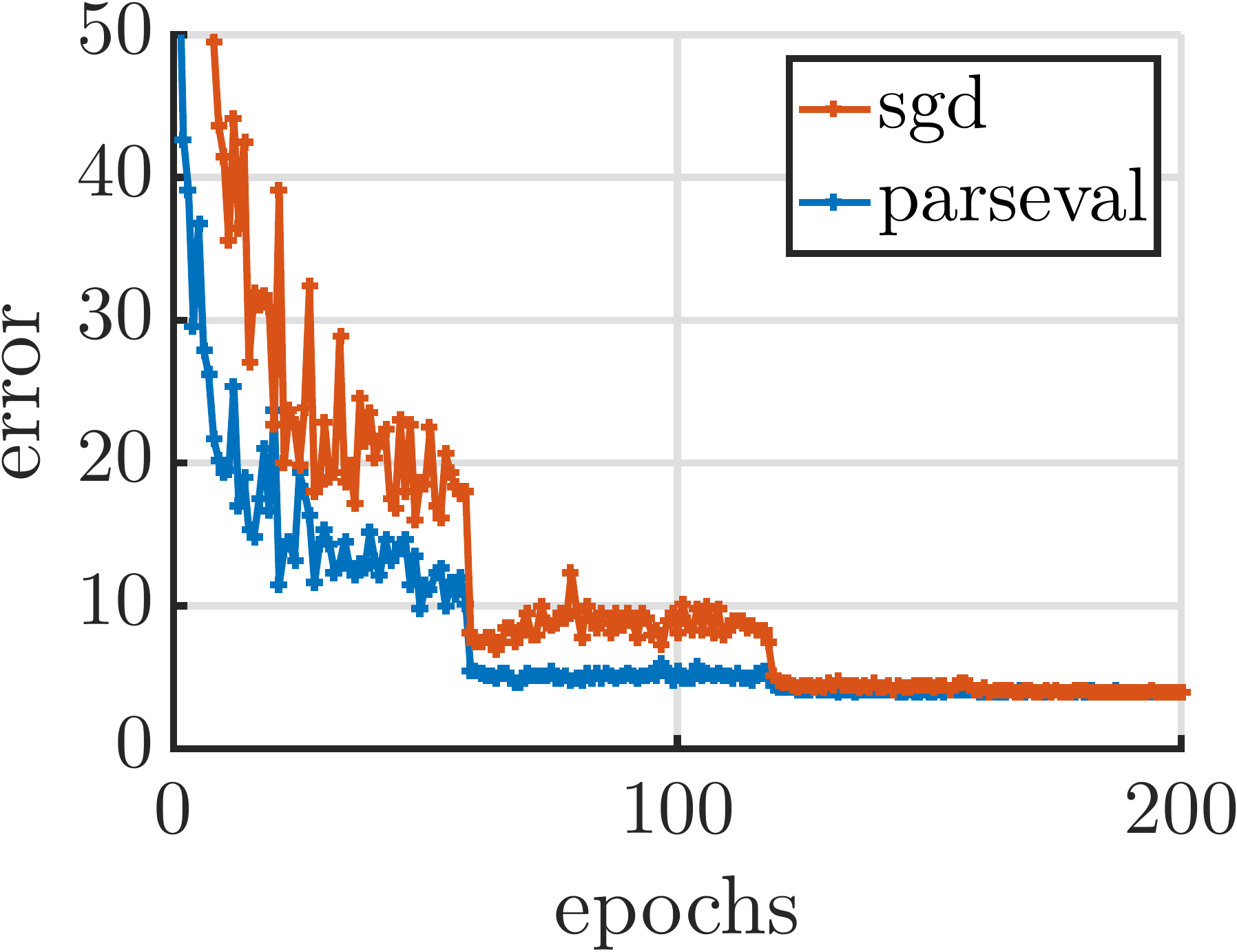}
    \includegraphics[width=0.49\linewidth]{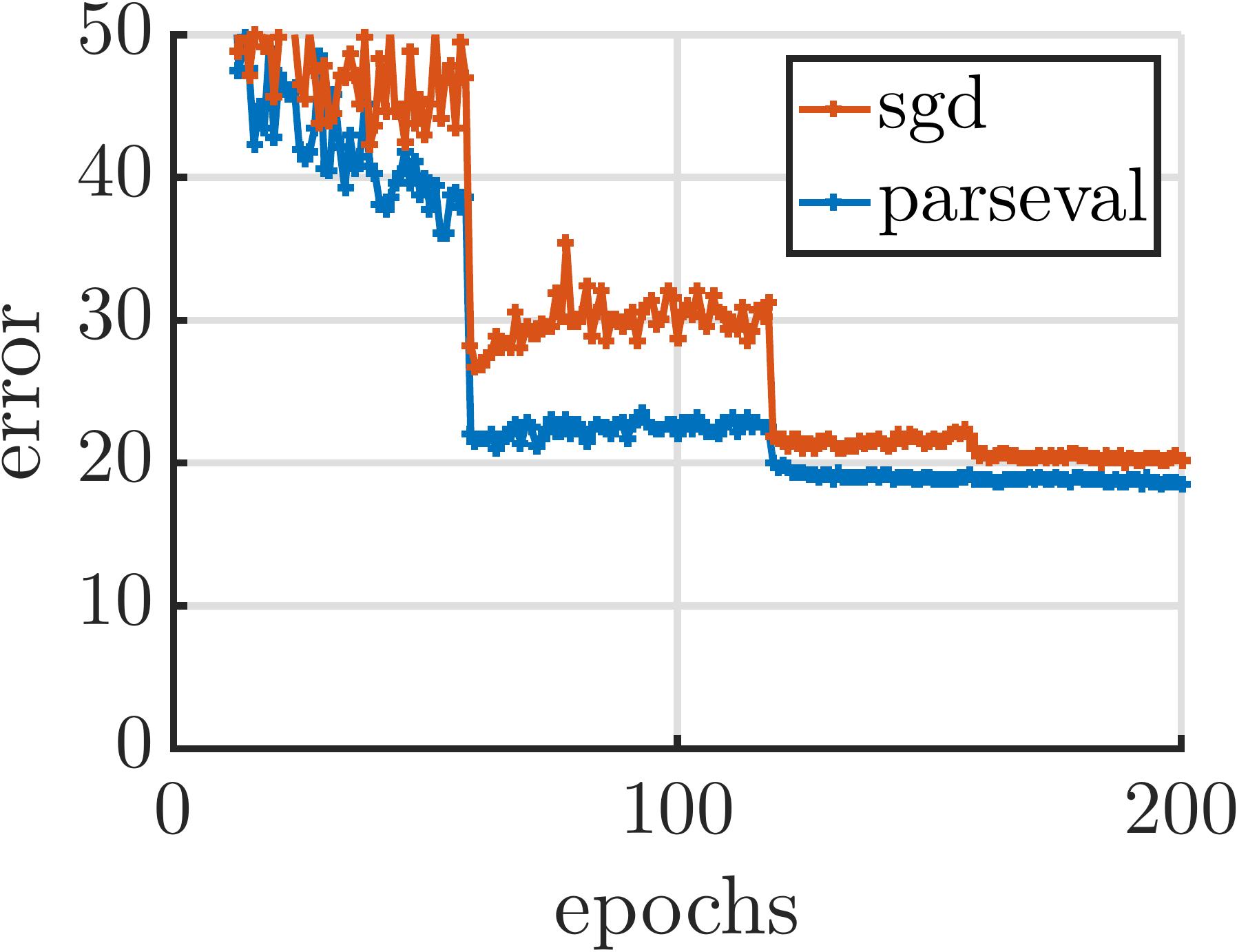}
    \caption{
Learning curves of Parseval wide resnets and Vanilla wide resnets on CIFAR-10 (right) and CIFAR-100 (left).
Parseval networks converge faster than their vanilla counterpart.
    }
    \label{fig:convergence}
    \vspace{-1.5em}
\end{figure}

Parseval networks converge significantly faster than vanilla networks trained with batch normalization and dropout as depicted by figure~\ref{fig:convergence}. Thanks to the orthogonalization step following each gradient update, the
weight matrices are well conditioned at each step during the optimization. We hypothesize this is the main explanation of this phenomenon. For convolutional networks (resnets), the faster convergence is not obtained at the expense of larger wall-time since the cost of the projection step is negligible compared to the total cost of the forward pass on modern GPU architecture thanks to the small size of the filters.

\section{Conclusion}

We introduced Parseval networks, a new approach for learning neural
networks that are intrinsically robust to adversarial noise.  We
proposed an algorithm that allows us to optimize the
model efficiently.  Empirical results on three classification datasets with fully connected and wide residual networks
illustrate the performance of our approach.  As
a byproduct of the regularization we propose, the model trains faster and makes a
better use of its capacity. Further investigation of this
phenomenon is left to future work.

\bibliography{parseval}
\bibliographystyle{icml2017}

\end{document}

%% file: macros.tex
\newcommand{\Nats}{\mathbb{N}}
\newcommand{\indic}[1]{\mathbbm{1}_{#1}}

\renewcommand{\expect}[2]{\mathop{\mathbb{E}}\limits_{#1}\big[#2\big]\!\!}
\newcommand{\calX}{{\cal X}}

\newcommand{\covnump}[1]{{\cal C}_p(\calX, #1)}
\newcommand{\coveps}{\gamma}

\newcommand{\wid}{k}
\newcommand{\Din}{d_{in}}
\newcommand{\Dout}{d_{out}}

\newcommand{\bx}{x}
\newcommand{\by}{y}
\newcommand{\bxprime}{\tilde{x}}
\newcommand{\bW}{W}

\newcommand{\uU}{U\xfunction}

\newcommand{\conv}{\ast}

\newcommand{\normt}[1]{\norm{#1}_2}
\newcommand{\norminf}[1]{\norm{#1}_{\infty}}
\newcommand{\normp}[1]{\norm{#1}_p}
\newcommand{\network}{g}
\newcommand{\cnet}{\hat{g}\xfunction}

\newcommand{\lipp}{\Lambda_p}
\newcommand{\liplossp}{\lambda_p}
\newcommand{\liplosst}{\lambda_2}
\newcommand{\liplossinf}{\lambda_\infty}

\newcommand{\lipgp}{\Lambda_p}
\newcommand{\lipgt}{\Lambda_2}
\newcommand{\lipginf}{\Lambda_{\infty}}

\newcommand{\bdc}{z}

\newcommand{\graph}{G}
\newcommand{\nodes}{{\cal N}}
\newcommand{\edges}{{\cal E}}
\newcommand{\nod}{n}
\newcommand{\nodind}[1]{^{(#1)}}
\newcommand{\layerfunc}{\phi}

\newcommand{\loss}{\ell}
\newcommand{\logsoftmax}{\ell}
\newcommand{\Loss}{L\xfunction}
\newcommand{\Lossadv}[3]{L_{adv}(#1, #2, #3)}

\newcommand{\Dx}{\Re^D}
\newcommand{\Dc}{\Re^{\nC}}
\newcommand{\Cset}{{\cal Y}}
\newcommand{\params}{W}
\newcommand{\Dparams}{{\cal W}}

\newcommand{\ccc}{{\bar{y}}}
\newcommand{\nC}{Y}

\newcommand{\adv}[1]{\tilde{#1}}
\newcommand{\perturb}[1]{\delta_{#1}}
\newcommand{\adveps}{\epsilon}